\documentclass[journal]{IEEEtran}
\usepackage{times}
\usepackage{epsfig}
\usepackage{graphicx}
\usepackage{subfigure}
\usepackage{amsmath}
\usepackage{amssymb}
\usepackage[ruled, linesnumbered]{algorithm2e}
\usepackage{color}
\usepackage{makecell}
\usepackage{multirow}
\usepackage{lineno,hyperref}

\ifCLASSINFOpdf
\else
\fi
\begin{document}
%
\title{Incremental Concept Learning via Online Generative Memory Recall}
%
%
%

\author{\IEEEauthorblockN{
		Huaiyu Li,
		Weiming Dong,~\IEEEmembership{Member,~IEEE},
		Bao-Gang Hu, ~\IEEEmembership{Senior Member,~IEEE}
	}

\thanks{Huaiyu Li was with National Laboratory of Pattern Recognition, Institute of Automation, Chinese Academy of Sciences,Beijing 100190, China and
also with University of Chinese Academy of Sciences, Beijing 100049, China(email: lihuaiyu2014@ia.ac.cn)}
\thanks{Weiming Dong and Bao-Gang Hu are with National Laboratory of Pattern Recognition, Institute of Automation, Chinese Academy of Sciences,Beijing 100190, China.(email: weiming.dong@ia.ac.cn; hubg@nlpr.ia.ac.cn)}
\thanks{Manuscript received July 5, 2019; revised XX XX, 2019.}}

%
%

\markboth{Journal of \LaTeX\ Class Files,~Vol.~XX, No.~XX, XX Month~2019}%
{Shell \MakeLowercase{\textit{et al.}}: Bare Demo of IEEEtran.cls for IEEE Journals}
%



\maketitle

\begin{abstract}
The ability to learn more and more concepts over time from incrementally arriving data is essential for the development of a life-long learning system. However, deep neural networks often suffer from forgetting previously learned concepts when continually learning new concepts, which is known as catastrophic forgetting problem. 
The main reason for catastrophic forgetting is that the past concept data is not available and neural weights are changed during incrementally learning new concepts.
In this paper, we propose a pseudo-rehearsal based class incremental learning approach to make neural networks capable of continually learning new concepts. 
We use a conditional generative adversarial network to consolidate old concepts memory and recall pseudo samples during learning new concepts and a balanced online memory recall strategy is to maximally maintain old memories. 
And we design a comprehensible incremental concept learning network as well as a concept contrastive loss to alleviate the magnitude of neural weights change. 
We evaluate the proposed approach on MNIST, Fashion-MNIST and SVHN datasets and compare with other rehearsal based approaches. The extensive experiments demonstrate the effectiveness of our approach.
\end{abstract}





%
\IEEEpeerreviewmaketitle

\section{Introduction}
The recent development of deep learning has achieved great success in a broad range of  computer vision tasks.
However, it is still far away from the purpose of artificial general intelligence(AGI). 
In the real world, artificial intelligence(AI) visual systems are usually exposed to dynamic environments where new visual concepts need to learn are emerging over time. 
In order to make the artificial visual systems move closer towards general intelligence, it is essential to make them possess the capability of \textit{continual} or \textit{lifelong learning} which means to continually learn over time by accommodating new knowledge while retaining previously learned experience\cite{2019CLreview}. 

In traditional learning scenario, successful deep neural networks(DNNs) are usually trained using stochastic gradient descent(SGD) optimization in batch mode where all training data are given at the same time and all classes are known in advance.
However, in continual learning scenarios, we have a neural network with the capability of recognizing some learned visual concepts and want to extend its capability of recognizing more visual concepts while giving new concept data. In practice, the most pragmatic way to achieve this goal is merging the data of learned concepts with the data of new concepts and retrain a deep neural network from scratch. 
Nevertheless, this methodology is pretty inefficient, since it requires large storage space and long retraining time. And in some cases, the previously learned concept data is never available.
If we directly train a trained neural network on new concept data, its performance of recognizing previously learned concepts  will significantly decreases. This phenomenon is known as \textit{catastrophic forgetting} or \textit{catastrophic interference} \cite{1989interference,1995cls} which refers to training a model using only new information without old information can lead to a severe performance degradation to old learned knowledge. 
Therefore, developing more flexible and intelligent methodologies to overcome catastrophic forgetting problem and achieve the continual learning goal is considerably significant and meaningful. Furthermore, gradually increasing the recognition ability of neural networks is a critical step to make learning systems closer to AGI.

From the perspective of human learning, they usually exhibit a strong ability to continually learn and accumulate knowledge throughout their lifespan. 
In particular, when they learn new knowledge, the new knowledge will not interfere with previous knowledge. Therefore, the catastrophic forgetting problem does not happen in human learning systems, which is a crucial characteristic for humans to behave intelligently. This is due to the special neurophysiological and biological mechanism of learning and memory of human brain which have been widely studied\cite{1987memory,2017neural}.
Among these studies, one important study is about the stability-plasticity dilemma\cite{2013stability} which refers to the extent of a learning system to be plastic in order to integrate novel information and stable in order not to severely interfere with consolidated knowledge. The lifelong learning in human brain is mediated by a rich set of neurophysiological principles that regulate the stability-plasticity balance of the different brain areas. 
Another important study is the complementary learning systems(CLS)\cite{1995cls} theory which illustrates the significant contribution of dual memory systems involving the hippocampus and the neocortex in learning and memory. It suggests that there are specialized mechanisms in the human cognitive system for protecting consolidated knowledge. The hippocampal system rapidly encodes recent experiences and exhibits short-term adapation. The long-term learning happens in neocortex systems through the activation synchronized with multiple replays of the encoded experience.
These studies of learning and memory of human brain have motivated many previous continual learning approaches\cite{2017DGR,2017DGDM,2018ICLwGAN}.
Because of the essential difference of learning mechanisms between biological neural networks and artificial neural networks, it is difficult to design one-to-one corresponding counterparts of hippocampal system and a neocortex system in artificial neural network learning systems. However, the fact that plasticity-stability balance and memory retrieval mechanism is indispensable ingredient in human learning, is convinced. 

According to different forms of deep neural networks, we can categorize the memory in artificial neural networks into three different types: implicit memory, explicit memory and recall memory.
The implicit memory refers to the learned neural weights of DNNs. Although a single connection weight may not show any concrete meanings, the whole learned neural weights in a neural network can exhibit the functionality for solving a specific task. The implicit memory is embedded in the neural weights and represents a latent form of learned knowledge.
The explicit memory refers to the external memory augmented in deep neural networks\cite{2014ntm,2015end,2018RNNAMU}. Compared with the implicit memory, the explicit memory is usually formed in definite structure such as a module of sequential vectors or matrix. This kind of memory can explicitly write and read to store and access information.
The recall memory refers to the concrete memory information of previous learned knowledge. This kind of memory can be represented by generative models\cite{2013vae,2014gan} which can recall pseudo learned information.

In order to make deep neural networks capable of continually learning to recognize more visual concepts, we should consider both the plasticity-stability balance during learning new concepts and how to utilize different types of memory to alleviate catastrophic forgetting problem in neural network models. 
Therefore, in this paper, we propose an incremental concept learning framework based on the characteristic of different types of artificial neural networks memory. It consists of two main components: an incremental concept learning network(ICLNet) and an incremental generative memory recall network(RecallNet).

The ICLNet is a remoulded feedforward neural network. It consists of a trainable feature extractorand a concept memory matrix which can be regared as implicit memory and explicit memory respectively. During incremental concept learning, we gradually increase new concept vectors in the concept memory matrix to make a neural network capable of recognize new concepts. 
Traditionally, the classifier of deep neural networks is usually a fully connected layer with softmax activation function and trained with cross entropy loss. However, training in this way will lead to large feature variance\cite{2018CosineNorm,2016centerloss}, which will result in larger weights variation and harm the implicit memory during continual learning.
Hence, we propose a contrastive concept loss for ICLNet which forces the learned features into a compact representation form. In this way, the learning capability of a neural network is not limited and the ability of preserving old concepts is improved. Hence, the plasticity and stability dilemma is balanced in when continually training neural networks.

The RecallNet aims to continually retrieve old concept memories while learning new concepts and incrementally consolidate new concept memories as well as old concepts.
We train conditional generative adversarial networks(cGANs) with least square loss as the RecallNet which is regarded as recall memory. By using cGANs, we can directly recall pseudo samples of old concepts. 
While training cGANs in incremental concept learning scenarios, we will also confront forgetting problems since existing generative models are still not satisfactory.
In order to maximally maintain old concept memories while incrementally training cGANs, we propose a balanced online recall strategy. In addition, this strategy can reduce the storage cost by not storing pseudo samples and also help other pseudo-rehearsal methods in continual learning.

We conduct a series of experiments on MNIST\cite{1998mnist}, Fashion-MNIST\cite{2017fashion} and SVHN\cite{2011svhn} datasets to illustrate the effectiveness of our proposed method. Compared with other pseudo-rehearsal\cite{2017DGR, 2018MRGAN}, we achieve the state of the art performance on the class incremental learning scenario.

\section{Related work}
In recent years, the research into continual learning is fairly compelling and has gradually gained widespread attention. 
The catastrophic forgetting effect is still a fundamental limitation for neural networks~\cite{1989interference} to continually learn new knowledge. 
It occurs when the data distribution of new tasks to learn is significantly different from previously observed data.
The main reason of the catastrophic forgetting phenomenon is that classical deep neural networks are trained with gradient-based optimization algorithms which aim at adapting neural weights in the network to fit current data distribution and loss function. 
In continual learning scenarios, the data distribution changes over time. 
When the neural network fits the new data distribution, the previously learned knowledge in the shared representational neural weights in the neural network will be overwritten by new information. Recently, many methods have been proposed for alleviating the catastrophic forgetting problem.

\subsection{Continual Learning Strategies}
From the perspective of learning strategies to alleviate catastrophic forgetting problem, recent methods can be roughly divided into four categories. 

\textbf{Regularization methods}.
The regularization methods alleviate catastrophic forgetting by imposing different kind of constraints while updating the neural weights\cite{2019CLreview}.  
One kind of regularization method is elastic weight consolidation(EWC) model \cite{2017EWC} and its following methods IMM \cite{2017IMM} and R-EWC \cite{2018REWC} which estimate the importance weights of a neural network for previous tasks and impose the constraint on the corresponding parameters while learning new tasks. The limitaions of this approach are that the shared parameters may contain conflicting constraints for different tasks and the importance weight matrix should be pre-estimated and stored.
Another well known regularization method is distillation. They generally use additional objective terms to regularize changes in the mapping function of a neural network. Learning without forgetting(LwF) \cite{2017LwF} enforce the predictions of a neural network for previously learned tasks to maintain similar while learning new tasks by using knowledge distillation \cite{2015distilling}. And the following work propose to use distillation on hidden activations \cite{2016less}, preserved old tasks data \cite{2017ICaRL,2018ICLwGAN} or additional dataset \cite{2019dmc} as regularization.
However, sequentially training with regularization terms usually limit the learning capability of neural networks in some extent. 

\textbf{Dynamic architecture methods}.
Dynamic architecture methods dynamically change the architecture of neural networks to accommodate new neural resources to learn new knowledge. The progressive networks \cite{2016progressivenn} approach proposed an architecture with explicit support for transfer learning across sequences of tasks. It expands the architecture by allocating new network branches with fixed capacity to learn new information.
Dynamically expanding network(DEN) \cite{2017DEN} can incrementally learn new tasks by increasing the number of trainable parameters. DEN is efficiently trained in an online
manner by performing selective retraining, and automatically determine how much neural networks capacity to expand. 
Fixed expansion layer(FEL) network\cite{2013ensembleCF} mitigates the forgetting by selectively update neural weights in the network rather than expanding the neural weights during continual learning.
Comparing with dynamic architecture methods, our approach only dynamically increase new concept vectors in concept memory matrix.

\textbf{Rehearsal based methods}.
The rehearsal based approaches select and store real representative samples from previous task training task which can maintain the information the past. ICaRL \cite{2017ICaRL} proposed a practical strategy for simulaneously learning a nearest-mean-of-exemplars classifier and a feature representation in the class incremental learning setting. ICaRL used \textit{herding} to create a fixed-sized represetation set of previous task data samples and the knowledge distillation to maintain past knowledge. Gradient episodic memory(GEM) \cite{2017GEM} and \cite{2018ICLwGAN} also require part of the old data. The main difference lies on how to select the representative exemplars and how to use the representative exemplars to help maintain knowledge of past tasks. The existing limitations of rehearsal based approaches is that it violates the data availability assumption and requires additional storage space. The continual learning performance will be influenced by the quality and quantity of stored representative examplars. 
Furthermore, in some real world applications, due to the privacy and legal concerns, real data may not be allowed to be stored for a long period of time\cite{2018ICLwGAN}.

\textbf{Pseudo-rehearsal based methods}.
The pseudo-rehearsal based methods \cite{2017DGR,2017DGDM,2018MRGAN,2018ICLwGAN} take inspiration from the CLS theory and make use of different promising generative models to learn the data distribution of previous tasks. During continually learning new tasks, the old task knowledge should be complemented by memory replay from learned generative models. The deep generative replay(DGR)\cite{2017DGR}, Memory Replay GANs \cite{2018MRGAN} and \cite{2018ICLwGAN} utilize variants of generative adversarial networks(GAN)\cite{2014gan} as generative memory. The deep generative dual memory network(DGM)\cite{2017DGDM} utilizes variational autoencoder(VAE)\cite{2013vae} as the long-term memory which stores information of all previously learned tasks. Our approach belongs to pseudo-rehearsal based method. In order to maintain information of generative models during sequentially learning new tasks, we propose a balanced online recall stratey which a general strategy and can be applied to other pseudo-rehearsal approaches.

\subsection{Continual Learning Scenarios}
Although promising results are reported by many methods, the experimental protocols and continual learning scenarios are different in their evaluation. And different continual learning scenarios have different learning difficulty levels. But they all confront catastrophic forgetting.
Therefore, from the perspective of learning scenarios, we can divide three distinct continual learning scenarios\cite{2018RTF}. 

\textbf{Task incremental learning scenario}. In this scenario, a new output-layer should be appended to the neural network while learning a new task. But during inference, the task identity should be provided to select which output-layer to use.

\textbf{Domain incremental learning scenario}. In this scenario, the same output-layer is shared while learning a new task. Therefore, for different learning tasks, they are constrained to have the same number of output units. And during inference, the task identity is also required. 

\textbf{Class incremental learning scenario}. In this scenario, the model can incrementally learn to recognize new classes and do not require to provide task identity during inference. There is only one incremental output-layer which continually increase the number of output units during learning new classes. In this paper, we refer the incremental concept learning as the same meaning with class incremental learning.

\begin{figure*}[t]
	\begin{center}
		\includegraphics[width=1.0\linewidth]{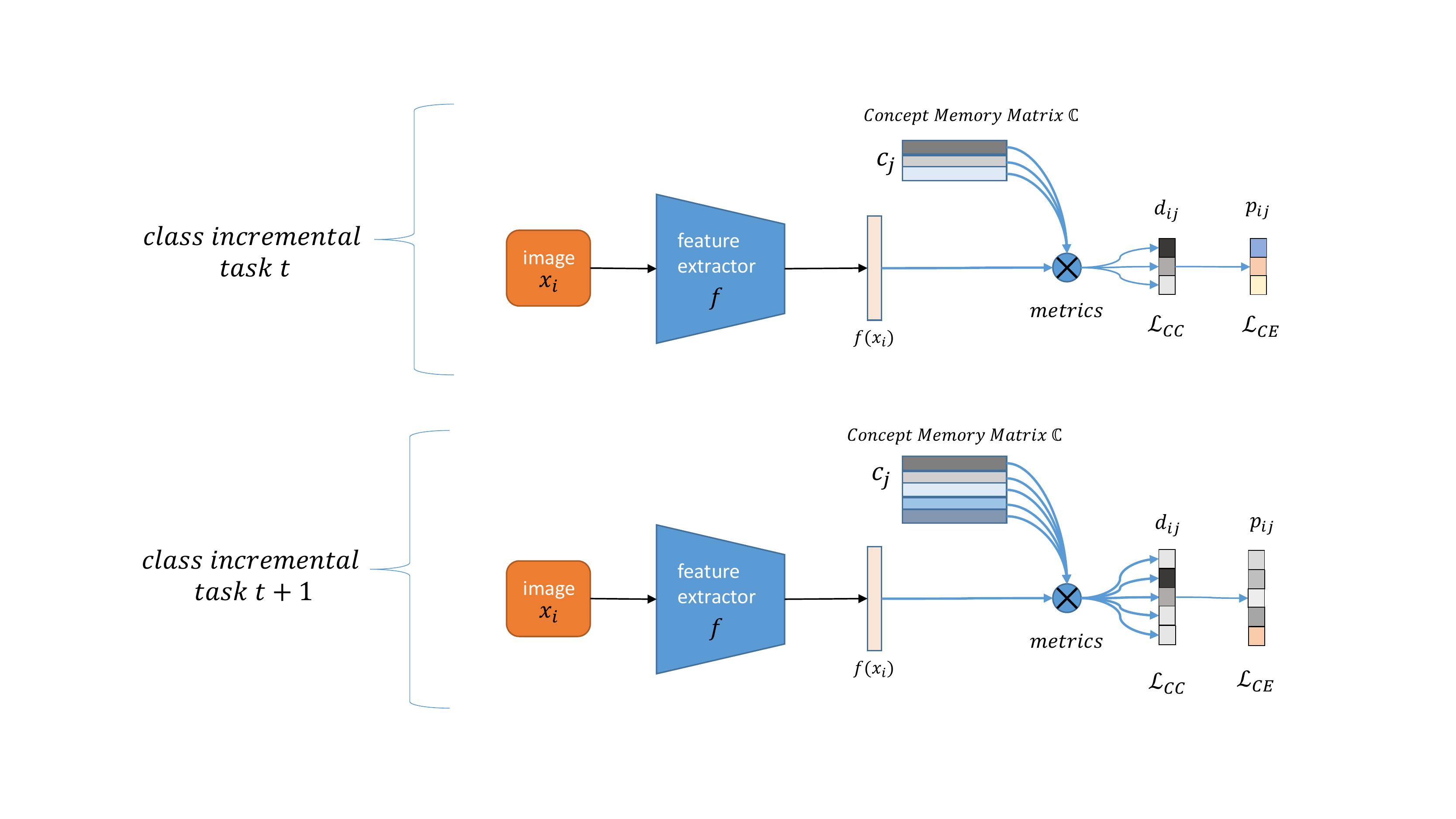}
	\end{center}
	\caption{The ICLNet architecture for class incremental learning tasks.}
	\label{fig:conceptnet}
\end{figure*}








\section{Method}
In this section, we describe the main components of our approach and explain how we design them. 
In Section~\ref{sec3.1} and \ref{sec3.2} , we describe the details of the incremental concept learning network(ICLNet) and the incremental generative memory recall network(RecallNet). 
In Section~\ref{sec3.3} , we describe the balance online recall strategy which helps to improve the continual learning ability in RecallNet.
At last, in Section~\ref{sec3.4} , we summarize the whole training procedure in an algorithm framework.

\subsection{Incremental Concept Learning Network}
\label{sec3.1}
From the perspective of training data in the continual learning scenarios, we have to face two major challenges:
(a). The new concept data has different distribution compared with old concept data. 
(b). And intact old concept data is usually not available while learning new concepts.
And from the perspective of training algorithm, we usually use stochastic gradient descent(SGD) algorithms\cite{2016sgdoverview} to update neural weights, which aims at adjusting the functionality of neural networks on current training data. And these two aspects are the main reasons for catastrophic forgetting problem happening in continually training deep neural networks. 
In order to balance the plasticity and stability in deep neural networks to allivate the catastrophic forgetting problem, we reinterpret the DNNs and utilize the implicit and explicit memory in DNNs.

The classical deep feedforward neural networks for classification tasks usually consists of a trainable feature extractor and a classifier. 
The output classifier is usually a fully connected layer which can produce raw value logits for each category. 
Followed by a softmax activation function layer, the network can predict the probabilities of each learned category for input samples. 
If the final fully connected layer classifier does not contain biases, it is equivalent to compute the inner product between input sample feature and each learnable vector in the weight matrix. 
Since the result of inner product is unbounded, if we train a classification neural network in this way, it will lead the feature extractor to produce features with large intra-class variance~\cite{2018CosineNorm,2016centerloss}, as shown in first row of figure~\ref{fig:2d}.
However, with large variance of the previous concept features, the neural networks weights become less stable during learning non-stationary data distribution in continual learning scenarios. 
The reasion is that the larger variance features make greater change in the neural weights while adapting to new concept data. Therefore, the old concept knowledge is tend to be forgotten more.
Previous methods ~\cite{2017LwF,2017ICaRL,2018ICLwGAN} use distillation loss to keep the ability of learner on previous concepts. But these methods require precomputed soft targets of replayed samples~\cite{2018ICLwGAN} or representative samples~\cite{2017ICaRL}, which increases the computational cost.

Because of previously mentioned concerns, we remould the neural network structure as ICLNet which is specifically designed for class incremental learning as shown in Figure~\ref{fig:conceptnet}.
We denote the ICLNet as $\phi^M$ with $M$ learned concepts. As illustrated in Figure \ref{fig:conceptnet}, the ICLNet consists of a trainable feature extractor $f: \chi \rightarrow \mathbb{R}^d$ and a dynamically growing concept memory matrix $\mathbb{C}$. The trainable feature extractor is an ordinary deep neural network such as multi-layer convolutional neural networks. And the neural weights of the feature extractor can be regarded as the implicit memory.
Each row vector in the concept memory matrix represents a concept prototype the network has learned. We can regard the memory matrix as the explicit memory. When the ICLNet need to learn new concepts, the concept memory matrix dynamically increase corresponding number of concept vectors. Unlike the external memory in NTM\cite{2014ntm} which learns differentiable controller to read and write memory, we directly update the concept memory matrix via gradients.

In our approach, we compute the distance $d_{ij}$ between sample features and each vector in concept memory matrix to predict the concept of samples. The closer the distance between features and concept vectors, the higher probability they belong to the corresponding concept. Specifically, we use Euclidean distance to measure the similarity. The probability of a sample $x_i$ belongs to concept $j$ is computed via a softmax function $\sigma$:
\begin{equation}
	p_{ij} = \frac{e^{-d_{ij}}} { \sum\limits_j e^{-d_{ij}}}
\end{equation}
where $d_{ij}=||f(x_i) - c_j||_2^2, j \in \{1,...,M\}, i \in \{1,...,N\}$ and $M$ is the number learned concepts of ICLNet and $N$ is the number of training samples. And we use cross entropy loss as our objective function:

\begin{equation}
\mathcal{L}_{CE} = - \frac{1}{N} \sum\limits^N_i \sum\limits^M_j y_{ij}log(p_{ij})
\end{equation}
where $y_{ij} \in \{0, 1\}$ is the groundtruth of sample $x_i$.

In the class incremental learning scenario, we need to train ICLNet with replayed samples combined with new concept data. However, directly train the ICLNet with $\mathcal{L}_{CE}$ on merged data will still lead to large intra-class feature variance in some learned concepts as shown in the second row of Figure~\ref{fig:2d}. 

In order to reduce the intra-class feature variance of learned concepts, we propose a concept contrastive loss to learn to represent the concept feature into a more compact form.
The concept contrastive loss aims at constraining the sample features to stay closer to corresponding concept vector but keep away from other concept vectors. The loss function is be formulated as following:
\begin{gather}
\mathcal{L}_{CC} =\frac{1}{N} \sum\limits^N_i \sum\limits^M_j y_{ij}*max(0, d_{ij}-\eta) \notag \\ 
+ (1-y_{ij})*min(0, d_{ij}-\beta*\eta)
\end{gather}
where $\eta$ is margin and $\beta$ is hyperparameter. 

Training the ICLNet with $\mathcal{L}_{CE}+\mathcal{L}_{CC}$ will make learned sample features near the corresponding concept vectors.
Hence, during incrementally learning new concepts in ICLNet, the learned old concept vectors in concept memory matrix are almost fixed while learning new concepts.
The neural weights of feature extractor are adjusted with smaller extent to keep old knowledge and learn new concepts.

\subsection{Incremental Generative Memory Recall Network}
\label{sec3.2}
In the concept incremental learning scenario, the intact old concept data is not available when learning new concepts. 
We propose to use a generative model to learn the distribution of old concepts as recall memory.
Currently, generative adversarial networks are the most promising generative models. In order to generate high quality samples and eliminate computational cost, we choose to use generative adversarial networks with auxiliary classifier(ACGAN) architecture which can directly generate samples for specified concepts. 
There are many candidate objective functions for training GANs, such as wasserstein generative adversarial networks(WGAN)\cite{2017wgan}, WGAN with gradient penalty(WGAN-GP)\cite{2017wgangp} and least square generative adversarial networks(LSGAN)~\cite{2017lsgan} etc.
Different from training cGANs on a single dataset, we require continually training cGANs on non-stationary datasets in continual learning scenarios. Therefore, the recall memory consolidation speed and stability are very important.
However, WGAN and WGAN-GP methods converge slowly and are hard to tune.
Therefore, we choose to use least square generative adversarial networks(LSGAN)~\cite{2017lsgan} which are more stable and faster on training ACGAN.

We denote the incremental generative memory recall Network(RecallNet) which can recall $M$ concepts as $\gamma_{M}$. 
When we are learning a new class incremental task(cit), we denote $\theta_{t} = (\theta_{t}^{G}, \theta_{t}^{D}, \theta_{t}^{CLS})$ as the parameter of generator, discriminator and auxillary classifier respectively within LS-ACGAN while learning the $t$th class incremental task. We alternatively train the generator and the discriminator with classifier by solving both adversarial game. The generator optimizes the following objective function: 
\begin{gather}
	\min_{\theta_t^G}  \mathcal{L}_{GAN}^{G}(\theta_t, S_t) + \mathcal{L}_{CLS}^{G}(\theta_t, S_t) \\
	\mathcal{L}_{GAN}^{G}(\theta_t, S_t) = -\mathbb{E}_{z \sim p_z, c \sim p_c }[(D_{\theta_t^D}(G_{\theta_t^G}(z, c))-1)^2] \\
	\mathcal{L}_{CLS}^{G}(\theta_t, S_t) =  -\mathbb{E}_{z \sim p_z, c \sim p_c }[y_{c} log(C_{\theta_{t}^{CLS}}(z, c)]
\end{gather}
where $S_t$ is the training set which combines recalled old concept samples and new concept samples, $p_c = \mathcal{U}\{1, M\}, p_z = \mathcal{N}(0, 1)$ are the sampling distributions.

Similarly, the discriminator and auxillary classifier optimize the following objective function:
\begin{gather}
	\min_{\theta_t^D, \theta_t^{CLS}}  \mathcal{L}_{GAN}^{D}(\theta_t, S_t) + \mathcal{L}_{CLS}^{D}(\theta_t, S_t) \\
	\mathcal{L}_{GAN}^{D}(\theta_t, S_t) = -\mathbb{E}_{(x,c) \sim S_t }[(D_{\theta_t^D}(x)-1)^2] \notag \\ 
										+ \mathbb{E}_{z \sim p_z, c \sim p_c}[D_{\theta_{t}^D}(G_{\theta_{t}^G}(z,c))^2] \\
	\mathcal{L}_{CLS}^{D}(\theta_t, S_t) =  -\mathbb{E}_{(x,c) \sim S_t }[y_{c} log(C_{\theta_{t}^{CLS}}(x))]
\end{gather}

\subsection{Balanced Online Recall}
\label{sec3.3}
Even though LS-ACGAN can stablize training and generate decent images, there are still many limitations in the exisiting techniques. Previous approaches\cite{2017DGR,2018ICLwGAN,2018MRGAN} leverage memory replay to create an extended dataset that contains both real new concept data and recalled pseudo samples for old concept data. However, creating an extended dataset will occupy more storage space and lose information within a limited amount of stored samples. If we continually train cGANs with the extended dataset, the information of long memory will gradually lose after a long number of class incremental tasks. Because a limited amount of recalled samples of old concepts can not cover the full data distribution.
If RecallNet fails to generate satisfactory images at one class incremental task learning stage, the following class incremental tasks learning will all be influenced.

Because of these mentioned problems, we propose a balanced online recall strategy. Since we usually train neural network with mini-batch stochastic gradient descent algorithm, we can directly recall old concept samples and combine with samples from new data to consist a training mini-batch. Therefore, we do not need to store the recalled old concept data.
Due to the concepts amount imbalance between the old memory and the new data, the conditional generator is not able learn to generate correct samples according to correct labels if we train RecallNet on the imbalanced image mini-batches.
And if we train ICLNet on the imbalanced batches, it also will lead to bias problems. This is confirmed by the experimental results.
Therefore, for each data batch, the amount of each concept samples should be proportional to the number of concepts in recall memory and new class incremental task.
In practice, for implementation simplicity, we proposed a balance sampling method which only requires to compute the number of samples for old concepts data and new concepts data. Assume that the training batch size is $B$ and the number of old concepts and new concepts are $M_o$ and $M_n$ respectively. We have
\begin{equation}
	B_o+B_n = B
\end{equation}
\begin{equation}
B_o/M_o = B_n/M_n
\end{equation}
where $B_o$ and $B_n$ is amount of old concept samples and new concept samples in a training batch respectively. By solving the simultaneous equations, we can get
\newcommand\round[1]{\left[#1\right]}
\begin{gather}
	B_o = \round{M_o*B/(M_o+M_n)} \notag \\
	B_n = \round{M_n*B/(M_o+M_n))}
\end{gather}
Sampling $B_o$ samples from RecallNet and $B_n$ samples from new data can consititue a class balance training batch. And we train both ICLNet and RecallNet using this strategy.

\subsection{Algorithm Framework}
\label{sec3.4}
\smallskip
\begin{algorithm}[t]
	\KwIn{New DataSet $D_{M_n}$, ICLNet $\phi_{M_o}$, RecallNet $\gamma_{M_o}$}
	\KwOut{ICLNet $\phi_{M_o+M_n}$, RecallNet $\hat{\gamma}_{M_o+M_n}$\\}
	Increase $M_n$ new concept vectors in concept memory matrix of ICLNet as $\phi_{M_o+M_n}$ \\
	\While{early stopping criterion not met}
	{
		$Tr_{batch} \leftarrow BalancedOnlineRecall(D_{M_n}, \gamma_{M_o})$ \\
		Update ICLNet $\phi_{M_o+M_n}$ by optimizing $\mathcal{L}_{CE}+\mathcal{L}_{CC}$ on $Tr_{batch}$
	}
	
	Initialize a new RecallNet $\hat{\gamma}_{M_o+M_n}$ \\
	\While{stopping criterion not met}
	{
		$Tr_{batch} \leftarrow BalancedOnlineRecall(D_{M_n}, \gamma_{M_o})$ \\
		Update RecallNet $\hat{\gamma}_{M_o+M_n}$ by optimize adversarial loss functions on $Tr_{batch}$
	}
	
	\caption{Training incremental concept learning network via incremental generative memory recall}
	\label{alg: overall}
\end{algorithm}
\smallskip
In Algorithm~\ref{alg: overall}, we present overall training procedure of our approach. When we need to learn a new class incremental task, we have a new dataset $D_{M_n}$ containing $M_n$ new concepts, the ICLNet $\phi_{M_o}$ which is capable of recognizing $M_o$ concepts and RecallNet $\gamma_{M_o}$ which can recall $M_o$ concepts samples. First, we increase $M_n$ new concept vectors in the concept memory matrix $\mathbb{C}$ of ICLNet.
Therefore, the ConceptNet $\phi_{M_o+M_n}$ can learn to distinguish $M_o+M_n$ concepts. Next, we use balanced online recall strategy to get balance training batches to train the ConceptNet. Finally, we initialize a new RecallNet $\hat{\gamma}_{M_o+M_n}$ which can learn to recall $M_o+M_n$ concepts samples. And then train the new RecallNet with balanced online recall strategy to consolidate the currently learned concepts memory.

\section{Experiment}
\begin{figure*}[htbp]
	\begin{center}
		\includegraphics[width=1.0\linewidth]{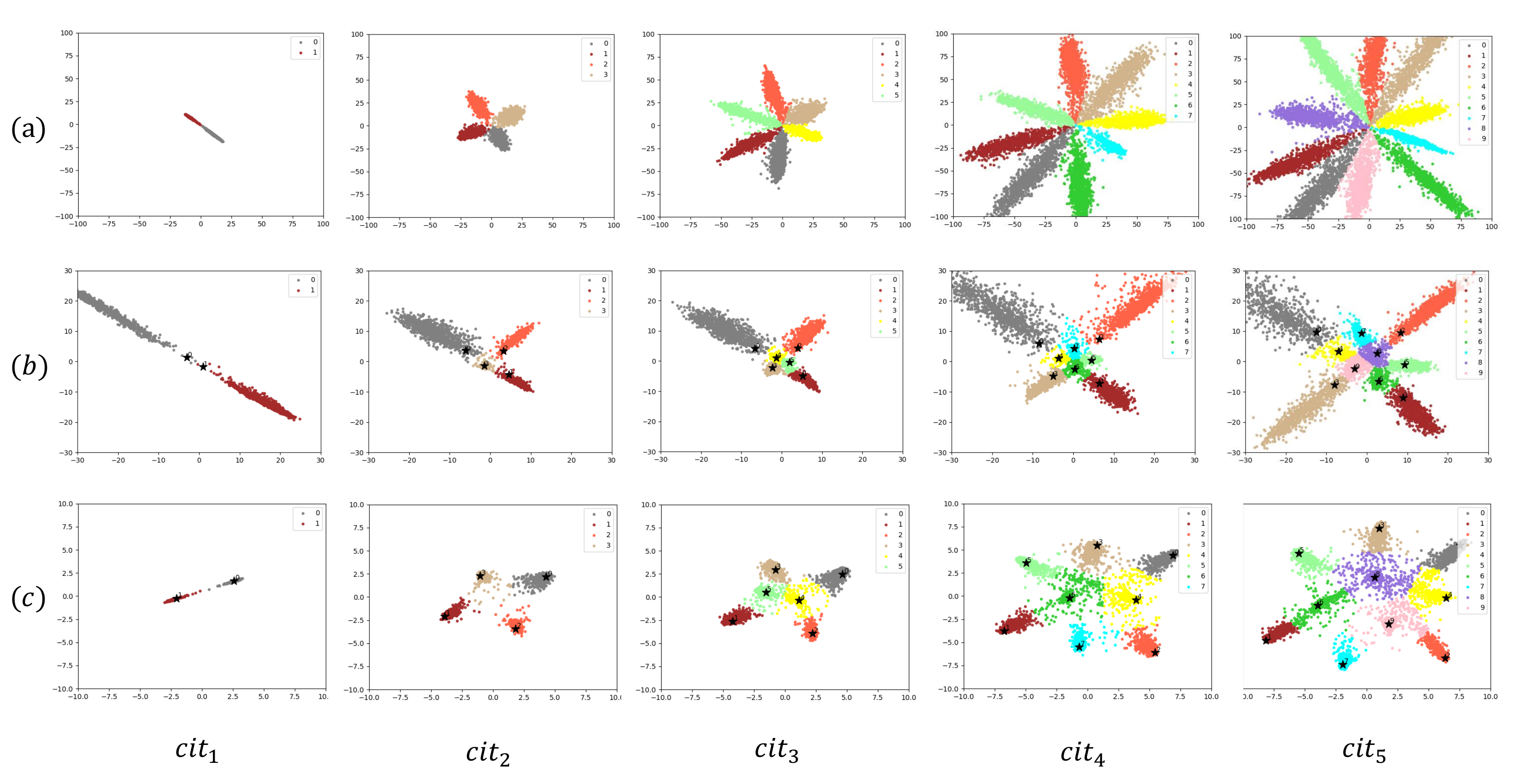}
	\end{center}
	\caption{The 2-D features visualization of currently learned classes during training on class incremental learning tasks. (a).Fully connected classifier trained with softmax cross entropy loss. (b). ConceptNet trained with softmax cross entropy loss. (c). ConceptNet trained with Softmax cross entropy loss and concepts contrastive loss. The black stars with labels in (b)(c) denotes the concept vectors of corresponding concepts.}
	\label{fig:2d}
\end{figure*}
In this section, we perform a variety of experiments to demonstrate the advantage of our approach. We present the implementation details in section \ref{sec4.1} . To illustrate the class incremental learning procedure, we present a feature visualization in section \ref{sec4.2} . We compare the continual learning performance with recently proposed related methods in section \ref{sec4.3} . In section \ref{sec4.4} , we perform ablation experiments to demonstrate the usefulness of key components of our approach. And show more experiments in section \ref{sec4.5} and \ref{sec4.6} .

\subsection{Implementation Details}
\label{sec4.1}
\textbf{Benchmark}: We use the \textit{average incremental accuracy} benchmark protocal suggested by~\cite{2017ICaRL} to evaluate the performance. It means, in class incremental scenarios, after each class incremental task(cit), the classifier is evaluated on the combined test set consisting of new and old class samples. We evaluate and compare on three different datasets. MNIST~\cite{1998mnist} consisting of images of handwritten digits and Fashion-MNIST~\cite{2017fashion} consisting of images of different fashion clothes are resized $32\times32$ pixels in our experiment. And SVHN~\cite{2011svhn} contains cropped digits of house numbers from real-world street images. Each of these datasets contains 10 classes. We can split each dataset into 5 class incremental tasks, each of which consists of 2 randomly picked classes. In our experiments, we directly pick classes for each task in the original label order which is $['0,1', '2, 3', '4, 5', '6, 7', '8, 9']$ for better understanding.

\textbf{Networks}: For MNIST and Fashion-MNIST, the feature extractor in ICLNet consists of 3 convolutional layers with batch normalization followed by a fully connected layer. And for SVHN, the feature extractor in ICLNet consists of 6 convolutional layers with batch normalization which is the structure provided by PyTorch\cite{2017pytorch} official examples. Since the memory consolidation is more challenging for Fashion-MNIST and SVHN which contain much more variability than MNIST, we use self-attention\cite{2018selfatten} and spectral normalization\cite{2018sn} techniques in the RecallNet. The detailed network implementations can be found in supplemental materials.

All experiemnts are implemented based on PyTorch\cite{2017pytorch} framework with Adam optimizer\cite{2014adam}, learning rate 1e-4 and batch size 64. Empirically,  we set the hyperparameter $\beta=4$ and $\eta=\sqrt{d}$ where $d$ is the dimension of concept vectors.

\begin{figure*}[t]
	\begin{center}
		\includegraphics[width=1.0\linewidth]{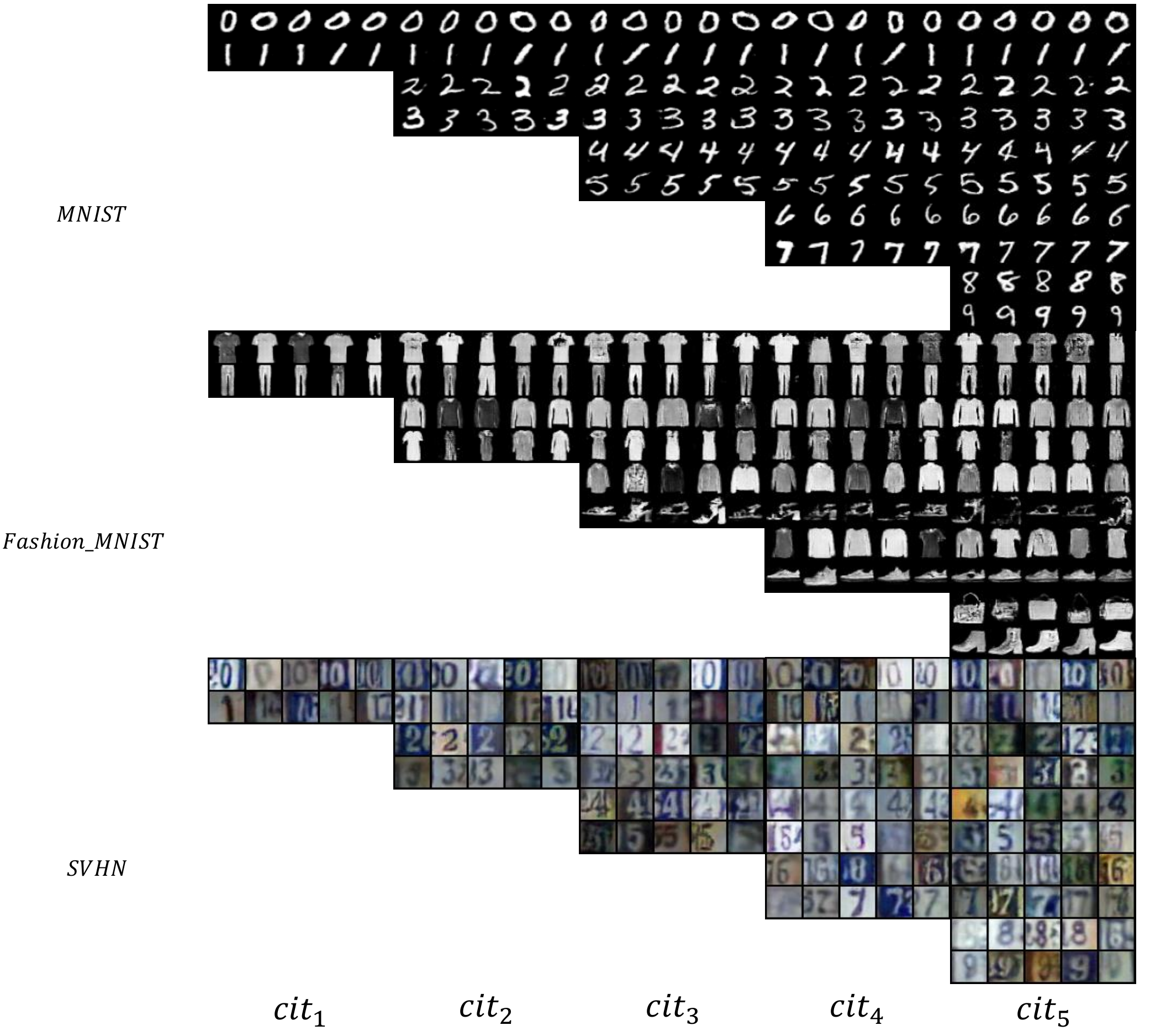}
	\end{center}
	\caption{The recalled samples from our RecallNet during learning 5 class incremental tasks on datasets MNIST, Fashion-MNIST and SVHN.}
	\label{fig:recall}
\end{figure*}

\begin{figure*}[htbp]
	\centering
	\includegraphics[width=1.0\linewidth]{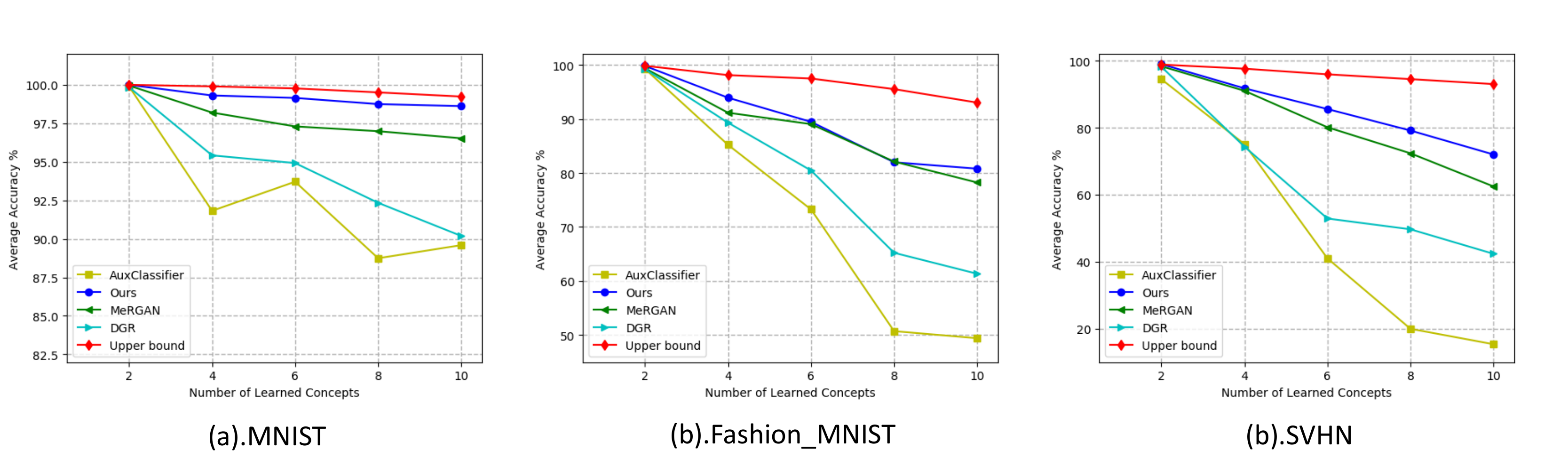}
	\caption{The average incremental accuracy for different methods during learning 5 class incremental tasks on datasets (a).MNIST, (b).Fashion-MNIST, (c).SVHN}
	\label{fig:comp}
\end{figure*}

\subsection{Visualization}
\label{sec4.2}
In this section, a toy example on MNIST dataset is presented. We reduce the output dimension of feature extractor to 2 as well as the dimension of concept vectors. So we can directly plot the features on 2-D surface for visualization. We show the class incremental learning procedure of different concepts on 5 class incremental tasks in Figure \ref{fig:2d} . Three conditions of different classifiers and losses are compared. 

In the first row, when we use traditional fully connected layer as classifier with softmax cross entropy loss, we can see that the numerical scale of feature and intra-class variance are gradually increasing as the continual learning progresses. In the continual learning scenario, when the neural networks incrementally learn new concepts, the large changes in feature scales can lead to large network weight changes. Thereby, it increases the impact of catastrophic forgetting problems. 
When train ICLNet only with softmax cross entropy loss, it reduces the feature scale but still has large intra-class feature variance as shown in the second row. 
To further reduce the intra-class feature variance, we train ICLNet with softmax cross entropy loss and concept contrastive loss as the results shown in the third row. We can observe that the features belong to the same concept concentrate to its corresponding concept vectors and keep away from other concept vectors. This phenomenon verify the purpose of concept contrastive loss. Furthermore, it also improves the continual learning performance and we show results in the ablation study \ref{sec4.4} .

\subsection{Results and Comparisons}
\label{sec4.3}

In this section, we perform both quantitative and qualitative evaluations on our approach. We compare with DGR\cite{2017DGR} and MeRGAN~\cite{2018MRGAN}, since we all belong to pseudorehearsal-based methods~\cite{1995rehearsal} and use GAN-based models for memory recall and consolidation. 
For fair comparisons, we reimplement them using the same network structures as ours in PyTorch with reference to their official code. 

The \textbf{DGR} method is originally evaluated on domain incremental learning problems. Therefore, in order to make original DGR approach fit the class incremental learning scenario, we make some reasonable modifications. Following \cite{2017DGR}, we separately train a feedforward neural network classifier and an unconditional generative adversarial network. The replayed images are stored and labeled with the most likely category predicted by the classifier trained on the previous task. However, this way may introduce some incorrect labels for continuous training and harm the classifier in continually learning new classes.

The \textbf{MeRGAN} method proposes to use a conditional GAN to replay past data and integrates the classifier into the discriminator of cGANs. It can help to avoid potential classification errors and biased sampling towards recent categories. They use replay alignment techniques to further prevent forgetting. But we compare with the version without replay alignment.

The \textbf{AuxClassifier} method refers to the auxilliary classifier in the AC-GAN of our model which also contains the ability of classification in the class incremental learning scenario. Comparing with MeRGAN, AuxClassifier does not inherit the AC-GAN parameter from previous AC-GAN. Hence, it lacks of the implicit memory of previous learned concepts.
But we use balanced online recall strategy rather than creating an extended dataset of current task and recalled samples to train.

The \textbf{Upper Bound} refers to directly train the classifier with cross entropy loss using the data of all class incremental tasks. We regard it as the performance upper bound for continual learning.

We show the comparison results on MNIST, Fahsion-MNIST and SVHN datasets in table \ref{tab: comp} . Our results outperform other pseudorehearsal methods. For MNIST, the average precison on 5 class incremental tasks of our approach is close to upper bound. However, for the Fashion-MNIST and SVHN, there is still a big gap with upper bound.

We visualize the samples of each learned concepts recalled from the RecallNet after consolidating memory for each class incremental tasks in Figure~\ref{fig:recall}. We can see that for MNIST, the RecallNet can generate clear and diverse images even after learning the last incremental task. However, for Fashion-MNIST and SVHN, although we can recognize the generated samples, they gradually lose details and variablity.

We draw the average incremental accuracy of different methods on different datasets during learning class incremental tasks in Figure~\ref{fig:comp}. We can see that our approach performs better than other pseudo-rehearsal approaches.


\begin{table}[t]
	\caption{
		The average incremental accuracy$(\%)$ comparison of different methods on datasets MNIST, Fashion-MNIST and SVHN.
	}  
	\label{tab: comp} 
	\begin{center}
		\begin{tabular}{|c|c|c|c|}
			\hline
			Method  & MNIST & Fashion-MNIST & SVHN	
			\\\hline  \hline
			AuxClassifier & 89.59  &  49.41	& 15.32 
			\\\hline 
			DGR	 &  90.19 &  61.34	&  52.34   
			\\\hline  
			MeRGAN	& 96.53  &  77.34	&  64.24    
			\\\hline
			Ours  	& \textbf{98.62}	& \textbf{80.79} & \textbf{72.20}   
			\\\hline 
			Upper bound & \textbf{99.24} & \textbf{93.02}  &  \textbf{96.13}	    
			\\\hline 
		\end{tabular}
	\end{center}
	\vspace{-6pt}
\end{table}

\subsection{Ablation Study}
\label{sec4.4}
\textbf{\begin{table*}[t]
		\centering
		\caption{
			The average incremental accuracy$(\%)$ comparison of using different key components in our approach.
		}  
		\label{tab: ablation} 
		\begin{tabular}{|c|c|c|c|c|c|}
			\hline
			\multicolumn{3}{|c|}{Variants} & \multicolumn{3}{c|}{Datasets} \\ \hline
			Net Type & Recall Strategy & Loss Function &  MNIST & Fashion-MNIST & SVHN \\ \hline \hline
			\textit{TradNet} & \textit{Balanced Online Recall } & $\mathcal{L}_{CE}$ & 96.45  &  74.98 & 66.15 \\ \hline %
			\textit{ICLNet} & \textit{Balanced Online Recall} & $\mathcal{L}_{CE}$ & 97.24  & 78.82  &  69.62 \\ \hline %
			\textit{ICLNet} & \textit{Balanced Online Recall} & $\mathcal{L}_{CE}+\mathcal{L}_{CC}$ &  \textbf{98.62} & \textbf{80.79} & \textbf{72.20} \\ \hline
			\textit{ICLNet} & \textit{Imbalanced Online Recall} & $\mathcal{L}_{CE}+\mathcal{L}_{CC}$ & 95.21  & 73.38 & 58.45 \\ \hline 
			\textit{ICLNet} & \textit{Offline Balanced Recall} & $\mathcal{L}_{CE}+\mathcal{L}_{CC}$ & 93.24  & 54.38 & 43.24 \\ \hline 
		\end{tabular}
\end{table*}}

We perform a series of ablation experiments to demonstrate that the key components of our approach indeed improve the performance of class incremental learning. The variants are as follows. 
(\romannumeral1). Which kind of classifier do we use, the traditional fully connected layer(\textit{TradNet}) or the ICLNet structure(\textit{ICLNet}). 
(\romannumeral2). Which kind of memory recall strategy do we use, balanced online recall (\textit{Balanced Online Recall}), imbalanced online recall (\textit{Imbalanced Online Recall}) or storing recalled samples with balanced sampling(\textit{Offline Balanced Recall}).
(\romannumeral3). Which kind of loss function do we apply, only softmax cross entropy loss($\mathcal{L}_{CE}$) or with concept contrastive loss($\mathcal{L}_{CE}+\mathcal{L}_{CC}$). 
The results in table \ref{tab: ablation} show that each of proposed component of our approach makes significant contribution to class incremental learning tasks. The ICLNet with concepts contrastive loss can reduce the feature scale and intra-class variance to help alleviate forgetting problem when recalled samples are not realistic enough.
Storing recalled sample will lose the information during continual learning of RecallNet and decrease the quality of recalled samples in the subsequent tasks. 
And without balanced sampling, online recall will fail to learn to generate samples corresponding to class labels when the number of learned concepts and new concepts are extremely imbalance. Hence, it results in failure in continual learning.

\subsection{Learn Dissimilar Concepts}
\label{sec4.5}
\begin{figure}[h]
	\begin{center}
		\includegraphics[width=1.0\linewidth]{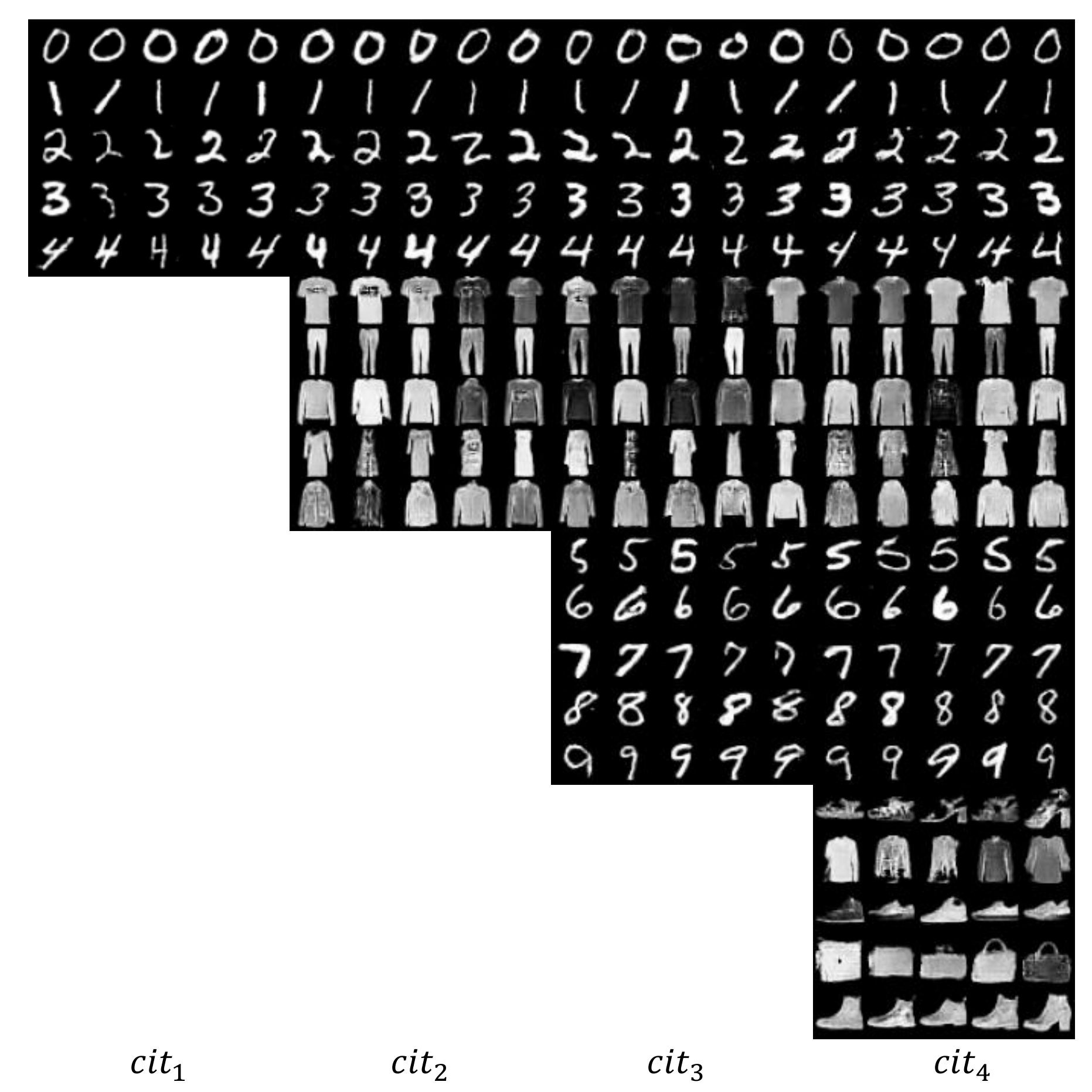} 
	\end{center}
	\caption{The recall samples from RecallNet after learning four class incremental tasks across dissimilar concepts.}
	\label{fig:acrossvis}
\end{figure}

\begin{figure}[h]
	\begin{center}
		\includegraphics[width=1.0\linewidth]{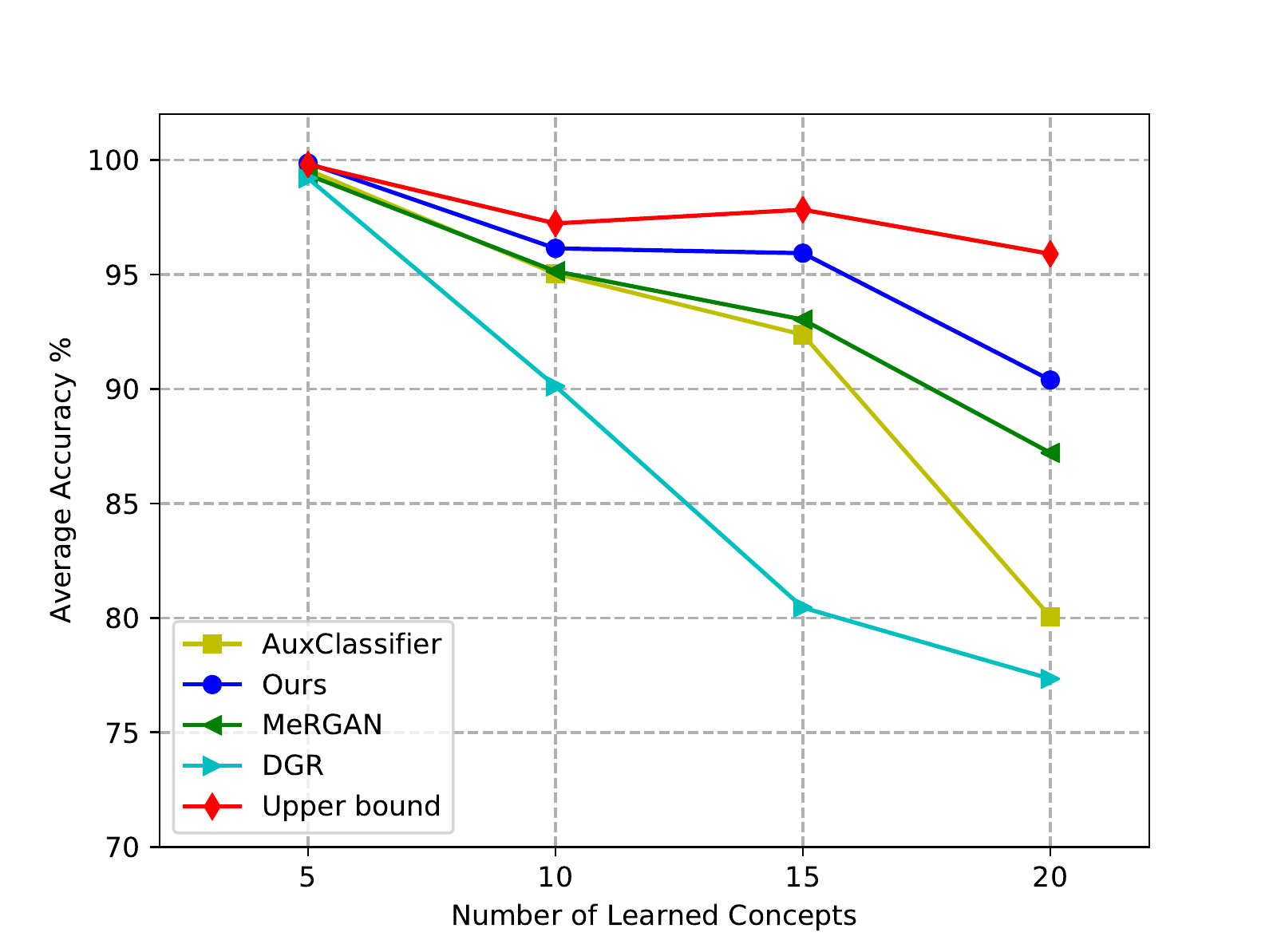} 
	\end{center}
	\caption{The average incremental accuracy for different methods during learning 4 class incremental tasks across dissimilar concepts.}
	\label{fig:acrossacc}
\end{figure}

We conduct a concept incremetal learning experiment across MNIST and Fashion-MNIST to show that our approach also works on continually learning dissimilar concepts from different datasets.
In this experiment, we split the whole training data of MNIST and Fashion-MNIST into four class incremental tasks which contain $['0, 1, 2, 3, 4', 'Tshirt, Trouser, Pullover, Dress, Coat',$ $'5, 6, 7, 8, 9', 'Sandal, Shirt, Sneaker, Bag, Ankle boot'$ concepts respectively. 
The average incremental accuracy of our approach have still surpass other pseudo-rehearsal based approach, DGR and MeRGAN, as shown in figure \ref{fig:acrossacc}.
We also visualize the recalled samples of each learned concepts after consolidating memory for each class incremental task in figure \ref{fig:acrossvis}. The quality incates that our approach is capable of maintain the memory of dissimilar concepts.

\subsection{Compare with Rehearsal-based Method}
\label{sec4.6}
\begin{figure}[h]
	\begin{center}
		\includegraphics[width=1.0\linewidth]{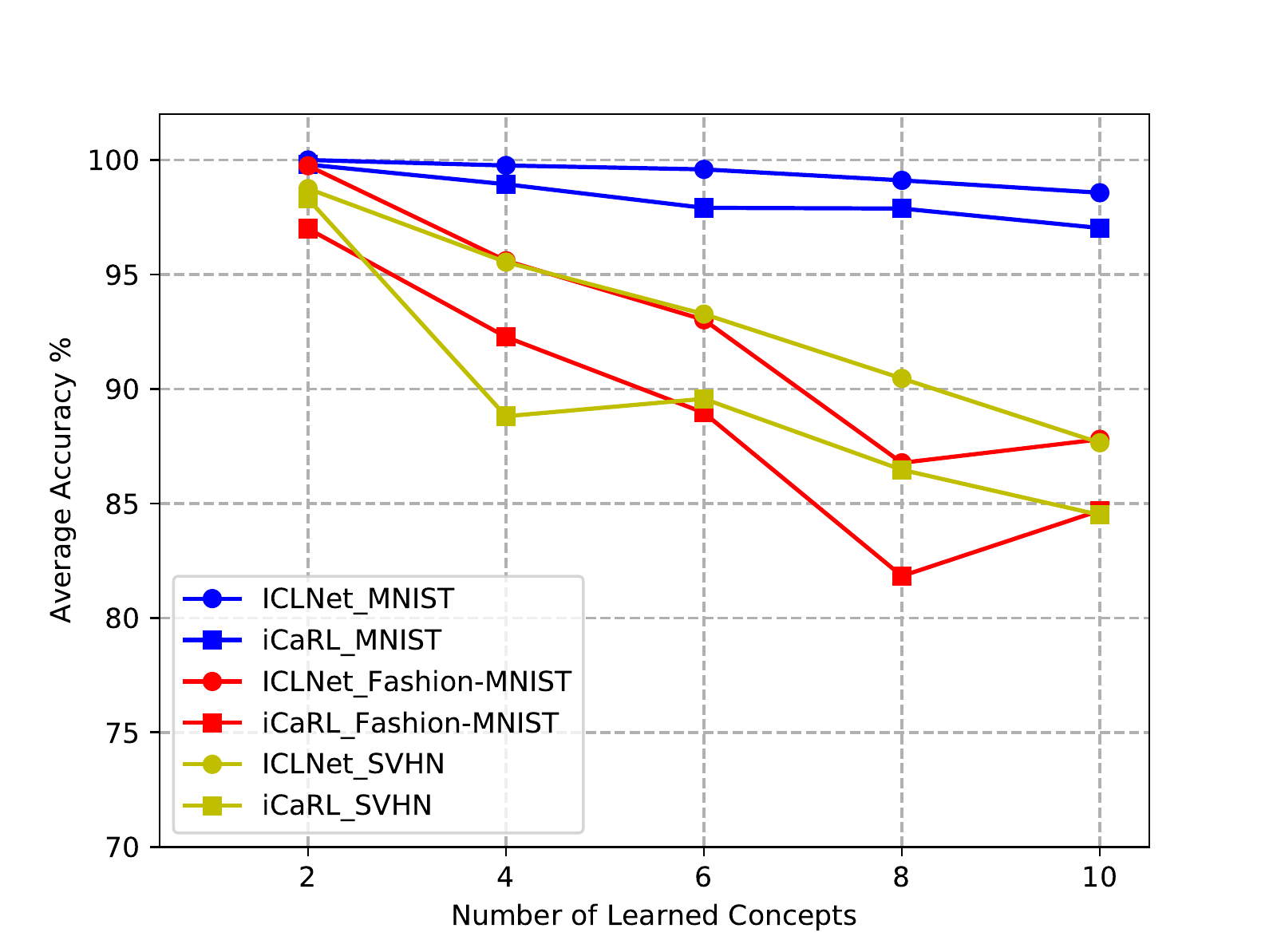} 
	\end{center}
	\caption{The average incremental accuracy for comparing ICLNet with iCaRL on rehearsal-based learning situation.}
	\label{fig:rehearsal_cmp}
\end{figure}

The ICLNet can also execute class incremental learning on the rehearsal-based situation which use selected representative sample set.
We compare with the well-known rehearsal-based method iCaRL which proposes to select representative samples for old class incremental tasks and use a distillation loss to train. In our reimplementation, we set the capacity of representative set $K=2000$ and use random exemplar selection rather than exemplar selection by herding since there is no substantial difference as claimed in \cite{2018revisitICL}. And we train our ICLNet with the same selected representative sample set for fair comparison. We conduct experiments on MNIST, Fashion-MNIST and SVHN and use the same the feature extractor in ICLNet and iCaRL. As shown in figure \ref{fig:rehearsal_cmp}, the ICLNet get better performance than iCaRL.

\section{Discussion}
In this paper, we propose a novel approach to solve catastrophic forgetting problem in class incremental learning scenario. In order to fit for class incremental learning scenario, we design a readily comprehensible network architecture named ICLNet which only need to increase the number of concept vectors in concept memory matrix while learning new concepts. 
To alleviate the catastrophic forgetting problem caused by large neural weights change, we propose a concept contrastive loss which help to concentrate the features to their corresponding concept vectors and keep away from other concept vectors. We can interpret the concept contrastive loss with ICLNet as a kind of regularization approach. 
Unlike EWC and IMM which regularize with precomputed importance weights and distillation methods \cite{2017LwF,2018ICLwGAN} which regularize  with soft targets of samples, we regularize with remoulded neural network structure and an loss function for continual learning.
We believe that redesigning computational architecture of neural networks will be a proper way to achieve efficient continual learning in the future.
On the other hand, we propose a balanced online recall strategy to make LS-ACGAN capable of incrementally consolidating generative memory with high quality in continual learning scenario.
Because of the advantage of the two proposed methods, we outperform other pseudo-rehearsal based approaches on MNIST, Fashion-MNIST and SVHN. And our ICLNet also performs better than well known rehearsal-based approach, iCaRL, in the same rehearsal scenario.

The disadvantage of our approach, which is also the disadvantage of other pseudo-rehearsal based approaches, is that we depend on 
the capability of the generative models used as recall memory. This may hinder applying pseudo-rehearsal based continual learning approaches to large scale dataset. However, the current development of generative models is very rapid. Excellent GANs like BigGAN \cite{2018BigGAN} and S$^3$GAN \cite{2019S3GAN} have been proposed and more progress will be made in the future. Hence, it is still promising to use GANs as recall memory. 


In the future, we aims to combine the ICLNet and RecallNet into an integrated model to improve computational efficiency and continual learning ability. Because the implicit and explicit memory in ICLNet should help consolidate recall memory in RecallNet.
And the information contained in RecallNet should also be used to help ICLNet continually learn new concepts.

\ifCLASSOPTIONcaptionsoff
  \newpage
\fi



\bibliographystyle{IEEEtran}
%

\bibliography{mybibfile}

\end{document}